\documentclass[runningheads]{llncs}
\usepackage{graphicx}
\usepackage{svg}
\usepackage{tabularx}
\usepackage{multirow}
\usepackage{subcaption}
\usepackage{amsmath}
\usepackage{graphicx}
\begin{document}
%
\title{Knowledge distillation with a class-aware loss for endoscopic disease detection}
%
%
%
\author{Pedro E. Chavarrias-Solano\inst{1}%
\thanks{}
\and Mansoor Ali-Teevno\inst{1} \and Gilberto Ochoa-Ruiz\inst{1} \and Sharib Ali\inst{2}}

\authorrunning{*}
%

\institute{Tecnologico de Monterrey, School of Engineering and Sciences, Mexico. \and  School of Computing, University of Leeds, Leeds, UK\\}

%
\maketitle              
\begin{abstract}
Prevalence of gastrointestinal (GI) cancer is growing alarmingly every year leading to a substantial increase in the mortality rate. Endoscopic detection is providing crucial diagnostic support, however, subtle lesions in upper and lower GI are quite hard to detect and cause considerable missed detection. In this work, we leverage deep learning to develop a framework to improve the localization of difficult to detect lesions and minimize the missed detection rate. We propose an end to end student-teacher learning setup where class probabilities of a trained teacher model on one class with larger dataset are used to penalize multi-class student network. Our model achieves higher performance in terms of mean average precision (mAP) on both endoscopic disease detection (EDD2020) challenge and Kvasir-SEG datasets. Additionally, we show that using such learning paradigm, our model is generalizable to unseen test set giving higher APs for clinically crucial neoplastic and polyp categories.  
\end{abstract}
\section{Introduction}
Gastrointestinal (GI) cancer accounts for 26\% of the global cancer incidence and 35\% of all cancer-related deaths, with colorectal, gastric and esophageal cancers being reported at 10.2\%, 5.7\% and 3.2\% rates, respectively~\cite{Arnold2020-ps}.  
Clinical endoscopy is thus critical for disease detection, diagnosis and risk categorisation of patients, as it allows a visual interpretation of mucosal changes. However, the detection of subtle lesions such as dysplasia (early neoplasia) can be difficult, thus leading to 11.3\% missed detection rates for neoplasia in the upper-GI and nearly 6\% missed polyps in the lower-GI surveillance~\cite{turshudzhyan2022lessons}. Artificial intelligence, and in particular, deep learning can help to improve the detection rates of hard to find lesions and minimise their missing rates, while assisting in their characterisation. 

Although several methods of this type have been developed in the literature, most focus has been on the polyp class with many datasets being publicly released and deep learning methods applied~\cite{shin2018automatic,urban2018yolo}. However, in reality, these methods cannot be used to find other inconspicuous lesions either on the same site or during a different endoscopic procedure. Positing this argument, the ``Endoscopic disease detection challenge 2020 (EDD2020)''~\cite{Ali_2020,ali2020endoscopy} released a dataset comprising of both upper-GI precancerous abnormalities such as Barrett's oesophagus, dysplasia, cancer and lower-GI anomalies including polyp and cancer. Motivated by this work, we aim to explore the opportunity this dataset has to offer to develop a unified deep learning framework for the entire GI tract. However, we also leverage other public datasets that have abundant polyp instances from the lower-GI surveillance (in our case Kvasir-SEG~\cite{jha2020kvasir}. 

The main rationale for our work is related to the difficulty of distinguishing between polyps, neoplasia and NBDE lesions in the entirety of the GI tract (see Fig.~\ref{fig:samples}). The EDD dataset is particularly challenging and it provides a good opportunity for testing methods that are robust and reliable in detecting such classes. As a matter of fact, the polyp class can be easily misclassified as neoplasia even for highly trained specialists, which can be problematic and thoroughly discussed above. Thus, herein we propose a method that leverages the teacher-student architecture used in knowledge distillation approaches as a way to mitigate these issues.
Our approach makes use of a teacher network (a Faster RCNN object detector) trained on Kvasir-SEG~\cite{jha2020kvasir} dataset to identify polyps with a higher certainty, whose layers are then frozen for the subsequent teaching process. In the next stage, this network is used to guide the training process of a student network (also a Faster RCNN) which is trained on a relatively small EDD2020 dataset~\cite{ali2020endoscopy} that contains polyps, neoplasia and NBDE lesion categories. 
%



The rest of the paper is organized as follows. In Section 2, we discuss the state-of-the-art methods in endoscopic disease detection and the limitations of the existing approaches. In Section 3, we discuss how we use endoscopic datasets in our experiments and we introduce our knowledge distillation framework and the proposed class-aware penalization loss function. In Section 4, we present the experimental setup and provide qualitative and quantitative results. Finally, Section 5 concludes the paper.
 \vspace{-5mm}
\section{Related work}
\begin{table}[t!]
\footnotesize
\begin{minipage}[b]{0.45\linewidth}
\centering
\caption{Dataset sample distribution between training and test for EDD2020 (revised) and Kvasir-SEG}
\label{table: Data Distribution}
\begin{tabular}{l|l|c|c|c}
\hline
\multicolumn{1}{c|}{\textbf{Dataset}} & \multicolumn{1}{c|}{\textbf{Type}} & \textbf{Train} & \textbf{Val} & \textbf{Test} \\ \hline
\textbf{EDD2020}                      & Images                             & 376            & 38           & 38            \\
                                      & NDBE                               & 239            & 28           & 19            \\
                                      & Neoplasia                          & 183            & 21           & 31            \\
                                      & Polyp                              & 172            & 24           & 32            \\ \hline
\textbf{Kvasir-SEG}                   & Images                             & 800            & 100          & 100           \\
                                      & Polyp                              & 858            & 111          & 102           \\ \hline
\end{tabular}
\end{minipage}\hfill
\begin{minipage}{0.48\linewidth}
\centering
\includegraphics[width=60mm]{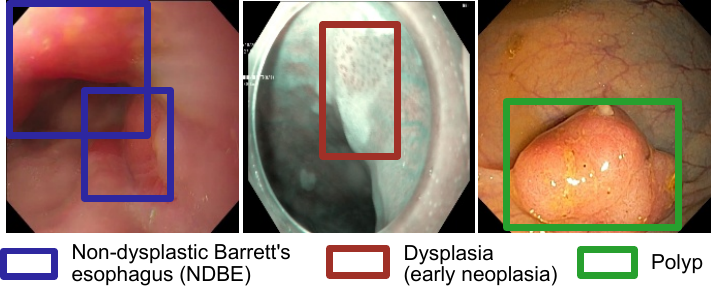}
\captionof{figure}{Representative samples of different classes in the EDD2020 dataset}
\label{fig:samples}
\end{minipage}
\end{table}
In the literature, several methods, including single-stage ~\cite{wang2018:segnet,urban2018yolo,horie2019} and two-stage \cite{yamada2019:2stage,shin2018automatic} networks, as well as anchor free architectures \cite{wang2019afp} 
have been used for the detection of different gastrointestinal diseases while maintaining real time performance. Some of the early approaches have exploited single stage detectors. For example, a SegNet \cite{segnet} based model was proposed by \cite{wang2018:segnet} to detect polyps during colonoscopy procedures. A real time polyp detection pipeline in colonoscopy based on YOLO was developed by Uraban \emph{et al.} \cite{urban2018yolo}. In another approach \cite{horie2019}, a single stage detector was leveraged for the early detection of multi-level esophageal cancer. Zhang \emph{et.al} \cite{zhang2018polyp} used pre-trained ResYolo architecture to extract features and then optimised output through a convolutional tracker. An SSD (ssdgpnet) based model exploiting a feature pyramid network was proposed in \cite{zhang2019real} for gastric polyp detection. More recently, two-stage detectors have shown improved accuracy and robustness, e.g., methods reported in EDD(2020)~\cite{Ali_2020}. A multi-stage detector uses a region proposal network to limit the search by generating candidates and then uses a classifier and bounding box regressor head to refine the search and produce final predictions. Among the two-stage networks, Faster RCNN \cite{FasterRCNN_2016} has been used as a base architecture in most detection pipelines owing to its improved precision. Yamada \emph{et al.} \cite{yamada2019:2stage} leveraged Faster RCNN with VGG as backone to detect hard samples of lesions normally ignored by colonoscopy procedures. They obtained encouraging results; however, the inference speed performance makes it unsuitable for real time examinations. In another study conducted by Shin \emph{et al.} \cite{shin2018automatic}, Faster RCNN with Inception and ResNet backbones were used to detect polyps. The two stage detectors have performed better over their single stage counterparts in EDD(2020) challenge~\cite{Ali_2020,krenzer2020}. In another approach \cite{jia2020automatic}, a two stage framework based on feature pyramid prediction was proposed for polyp identification from colonoscopy images. Contrast enhanced colonoscopy images were fed to an improved Faster RCNN architecture to boost polyp detection performance by authors in \cite{chen2021self}. {Deep networks} have been successful in yielding state-of-the-art results, but they require training huge models on large datasets for obtaining that performance. This makes them quite computationally expensive. An alternative approach is to use knowledge distillation for developing smaller and efficient models. This involves a trained teacher network to gradually transfer knowledge to a compact student network~\cite{Gou_2020}. Leveraging this idea, a teacher-student mutual learning pipeline with single teacher, multi-student framework was proposed in \cite{niyaz2022KD} for the polyp detection using hyper-Kvasir dataset. Henrik \emph{et al.}~\cite{gjestang2021:KD} used a knowledge distillation framework to perform a semi-supervised polyp classification showing a few points improvement in dice score over previous methods. Self-distillation was used for the early diagnosis of neoplasia in Barrett’s esophagus \cite{hou2021early}. Knowledge distillation along with video temporal context was used to to implement a real-time polyp detection framework by Li \emph{et al.} \cite{li2020real}.
Motivated by these works, we propose to use a fully trained (teacher) network on one class and use the predicted class probability to penalise the multi-class (student) learner.


%
%
\section{Materials and method}
\subsection{Datasets}
The EDD2020 dataset~\cite{Ali_2020} includes multi-center, multi-modal, multi-organ, multi-disease images comprising of both upper and lower GI. The dataset is encompassed by 386 video frames with 749 annotations that were collected using three different endoscopic modalities (white light, narrow-band imaging, and chromoendoscopy) at four different clinical centers during the screening of four different gastrointestinal organs. The detection and segmentation task is related to five disease categories: non-dysplastic Barrett's (286), suspicious (98), high-grade dysplasia (80), cancer (57), and polyp (228). Barrett's (NDBE) and polyp classes have more recurrences, while cancer, high-grade dysplasia (HGD), and suspicious have less samples. Some representative samples of the dataset are shown in Figure \ref{fig:samples}. 
EDD2020 dataset has large class imbalance so we merged cancer, HGD and suspicious classes into one class as ``neoplasia''.
%
The final data distribution is shown in Table \ref{table: Data Distribution}. We have made use of an additional dataset, the Kvasir-SEG~\cite{jha2020kvasir} that is widely used to develop and compare methods built for polyp detection and segmentation in the colon and rectum. This dataset contains 1,000 polyp images with their corresponding bounding boxes and segmentation masks. The second row in Table~\ref{table: Data Distribution} also summarizes the sample sizes of this dataset.

\begin{figure}[t!]
    \centering
    \includegraphics[width=\textwidth]{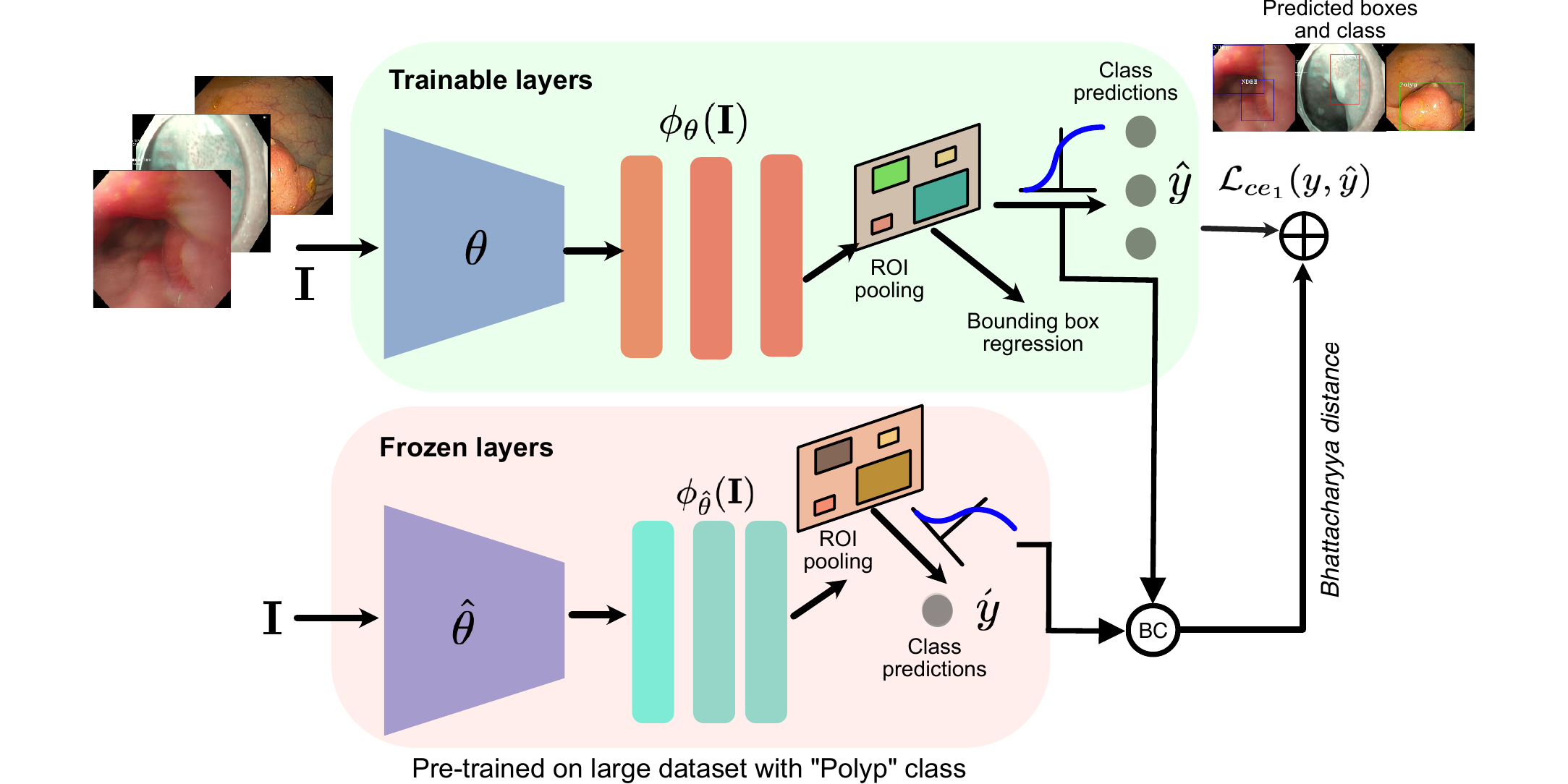}
    \caption{Featured-based knowledge distillation (proposed framework). The student (trainable layers) and teacher (frozen layers) networks are both implemented using a Faster R-CNN architecture. Here, the teacher model layers are pre-trained on a larger polyp dataset after which they are frozen. The feature vectors of class wise probability distributions are extracted from intermediate layers of both models. The Bhattacharyya distance is then computed between these probability distributions of both networks for each class separately, with varying weights and is added as a penalization term to the student classification loss.}
    \label{fig:blockdiagram}
\end{figure}
\subsection{Proposed knowledge-distillation framework}
Figure \ref{fig:blockdiagram} represents the block diagram of our proposed network. Here, the class probability distribution vectors of both networks are used to compute  the Bhattacharyya distance for each class, which is then used as a penalization term together with the standard cross-entropy loss for optimizing the parameters of the student model. The components of our framework are described as follows:

\paragraph{\textbf{Faster R-CNN}} an unified network for object detection that is composed of two main modules: a deep fully convolutional network that proposes regions (\textit{aka} Region Proposal Network (RPN)), and the Fast R-CNN detector that uses the pooled areas from the proposed regions \cite{FasterRCNN_2016} for classification and bounding box regression. 
The RPN takes the output of the last shared convolutional layer as input and produces a set of rectangular object proposals, each with an objectness score. To generate region proposals, the RPN takes as input a $n\times n$ window of the input feature map and maps it to a lower-dimensional feature. It is then fed into two fully-connected layers: a box-regression layer and a box-classification layer. At each window, the algorithm predicts multiple region proposals which are parameterized relative to the reference boxes, known as anchors centered at their corresponding window. This approach allows sharing features without extra computations for for  capturing different size objects in a multi-scale fashion \cite{FasterRCNN_2016}.

\paragraph{\textbf{Class-aware loss function.}}
We propose a class-aware loss to boost the performance of the object detection model. Our implementation follows a feature-based knowledge distillation approach \cite{Gou_2020}, in which we use a Faster R-CNN teacher network pre-trained in the Kvasir-SEG dataset for detecting polyps with a high confidence. The student network, also a Faster R-CNN architecture, compares the probability maps generated by its feature extraction network and those of the teacher network class probabilities. The rationale for this process is to implement a manner for penalizing the student network each time a polyp sample is classified incorrectly by the student network, but correctly detected by the teacher. Severe neoplastic cases that can confuse with polyp class are penalized with higher weights as that those with polyps. Similarly, for the NDBE cases, the penalisation factor ($\lambda$) is set relatively lower as these are slightly different disease cases but can confuse with early neoplastic changes. These $\lambda$ weights for each disease cases were empirically set.

Bhattacharyya distance ($\mathit{B}_{d}$) is computed between each prediction logits of the student ($p_s$) and teacher networks ($p_t$), as shown in Eqs. (\ref{eq1}-\ref{eq3}). A set of experimentally determined fixed weighting factors $\lambda$ are assigned to each distance. The sum of all distances is then computed and normalized as in Eq.~(\ref{eq4}). The normalization factor is determined by adding weights determined for each class ($\sum\lambda_{class} = \lambda_{ndbe} + \lambda_{neoplasia}+ \lambda_{polyp}$ ). 
%

\vspace{-5mm}

\begin{align}
\label{eq1}
B_{d-ndbe}(\mathbf{p}_{s_{ndbe}},\mathbf{p}_{t})= - \lambda_{ndbe}\cdot \ln(\sum_{i=1}^{n}{\sqrt{{p}^{i}_{s_{ndbe}},p^{i}_{t}}})\\
\label{eq2}
B_{d-neoplasia}(\mathbf{p}_{s_{neoplasia}},\mathbf{p}_{t})= - \lambda_{neoplasia} \cdot \ln(\sum_{i=1}^{n}{\sqrt{{p}^{i}_{s_{neoplasia}}, {p}^{i}_{t}}})\\
\label{eq3}
B_{d-polyp}(\mathbf{p}_{s_{polyp}},\mathbf{p}_{t})= - \lambda_{polyp} \cdot \ln(\sum_{i=1}^{n}{\sqrt{p^{i}_{s_{polyp}},p^{i}_{t}}})
\end{align}


The resultant distance $D$ is then added to the cross-entropy loss $\mathcal{L}_{ce}$ used to update the student weights. The weights are hyper-parameters of our model that were found empirically through a grid search. 

\begin{equation}
\label{eq4}
D = \frac{\sum{B_{d-class}}}{\sum{\lambda_{class}}}
\end{equation}

\paragraph{\textbf{Data augmentation techniques.}}
Two types of data augmentation techniques were applied to both datasets: geometric and photometric. For the geometric data augmentations we refer to 90 degrees random rotation, horizontal flip, vertical flip, and center crop. While, photometric image augmentations entail to transformations such as blur, image equalization, as well as changes in contrast, brightness, darkness, sharpness, hue, and saturation of the images. All augmentations are only applied to training set. 

\section{Experiments and results}
\subsection{Experimental setup and evaluation metrics}
All models were trained on an NVIDIA GeForce GTX 1650 GPU. Images were resized to $225 \times 225$ pixels and the following training, validation and testing splits is performed on both datasets with an 80:10:10 split. The data augmentation techniques described in Section 3.2 are used in all our models and we use $k$-fold with $k=3$ cross validation for unbiased comparison between models. All the Faster R-CNN sub-modules in our teacher-student framework  make use of a ResNet-50 backbone and were run with similar hyper-parameters that include a learning rate of 1e-3, momentum of 0.9, and a weight decay of 5e-4. Stochastic Gradient Descent (SGD) was used as an optimizer. 


We use standard computer vision metrics for detection that includes average precision (AP) computed for each class and mean average precision (mAP) over all three classes. Each of these metrics are evaluated at different IoU thresholds and are usually represented as $\text{AP}_{\text{IoU threshold}}$. We present results for IoU thresholds of 25, 50 and 25:75 (meaning averaged value for IoU threshold between 25 and 75 with spacing of 5). Usually, AP$_{50}$ is considered to have indicative results with good localization and classification score.

\vspace{-2mm}
\subsection{Results}
\vspace{-1mm}
We first trained Faster R-CNN using the Kvasir-SEG dataset. Then, we applied an informed class-aware knowledge distillation to the student network by applying the frozen pre-trained model on Kvasir-SEG (larger dataset on with polyp class) to improve the performance of the model on the EDD2020 dataset by leveraging the class-aware knowledge of this network. 
\vspace{-3mm}
\subsubsection{Quantitative results}
Table~\ref{table: KvasirSEG} present the results on Kvasir-SEG test split showing an improvement (nearly 3\% more compared to other augmentation) with our proposed combined geometric and photometric augmentation strategy. Table~\ref{table: KD results} demonstrate that the use of class-aware loss in our proposed knowledge distillation concept improves the mAP at different IoU thresholds for hold-out test samples (3.7\% on mAP$_{25}$ and 2.2\% on mAP$_{50}$). Similarly, for the class wise APs, a boost in NDBE and polyp classes can be observed with over 6\% and 4\%, respectively. Similarly, Table~\ref{table: Results unseen} demonstrate that the proposed approach outperformed the Faster R-CNN approach at all mAPs. Also, for the neoplasia class and polyp class, our approach showed an improvement of 3.3\% and nearly 13\%, respectively. 



\begin{table}[!t]
\centering
\caption{\textbf{Evaluation on Kvasir-SEG dataset using Faster R-CNN.} Evaluation results using mAP$_{25}$, mAP$_{50}$, mAP$_{75}$, and mAP$_{25:75}$ evaluation metrics}
\label{table: KvasirSEG}
\begin{tabular}{lllcccc}
\hline
\multicolumn{1}{l|}{\textbf{Method}}                                                                     & \multicolumn{1}{l|}{\textbf{Augmentation}}                                           & \multicolumn{1}{l|}{\textbf{Epochs}} & \multicolumn{1}{c|}{\textbf{mAP$_{25}$}} & \multicolumn{1}{c|}{\textbf{mAP$_{50}$}} & \multicolumn{1}{c|}{\textbf{mAP$_{75}$}} & \textbf{mAP$_{25:75}$} \\ \hline
\multicolumn{1}{l|}{\multirow{6}{*}{\begin{tabular}[c]{@{}l@{}}Faster R-CNN\\ (ResNet-50)\end{tabular}}} & \multicolumn{1}{l|}{None}                                                            & \multicolumn{1}{l|}{60}              & \multicolumn{1}{c|}{95.3}           & \multicolumn{1}{c|}{90.4}           & \multicolumn{1}{c|}{68.8}           & 86.6              \\ \cline{2-7} 
\multicolumn{1}{l|}{}                                                                                    & \multicolumn{6}{c}{\textbf{3-Fold Cross Validation done at mAP$_{50}$}}                                                                                                                                                                                                \\ \cline{2-7} 
\multicolumn{1}{l|}{}                                                                                    & \multicolumn{1}{l|}{}                                                                & \multicolumn{1}{l|}{}                & \multicolumn{1}{c|}{\textbf{K1}}    & \multicolumn{1}{c|}{\textbf{K2}}    & \multicolumn{1}{c|}{\textbf{K3}}    & \textbf{Average}  \\ \cline{2-7} 
\multicolumn{1}{l|}{}                                                                                    & \multicolumn{1}{l|}{Geometric}                                                       & \multicolumn{1}{l|}{20}              & \multicolumn{1}{c|}{82.4}           & \multicolumn{1}{c|}{85.5}           & \multicolumn{1}{c|}{85.1}           & 84.3              \\ \cline{2-7} 
\multicolumn{1}{l|}{}                                                                                    & \multicolumn{1}{l|}{Photometric}                                                     & \multicolumn{1}{l|}{20}              & \multicolumn{1}{c|}{87.3}           & \multicolumn{1}{c|}{83.6}           & \multicolumn{1}{c|}{86.5}           & 85.8              \\ \cline{2-7} 
\multicolumn{1}{l|}{}                                                                                    & \multicolumn{1}{l|}{\begin{tabular}[c]{@{}l@{}}Geometric$+$photometric\end{tabular}} & \multicolumn{1}{l|}{20}              & \multicolumn{1}{c|}{\textbf{87.6}}  & \multicolumn{1}{c|}{\textbf{85.6}}  & \multicolumn{1}{c|}{\textbf{88.1}}  & \textbf{87.1}     \\ \hline
\end{tabular}
\end{table}


\begin{table}[t!]
\centering
\caption{\textbf{Evaluation on 10\% held-out test data.} Faster R-CNN ( ResNet-50) without (SOTA) and with informed polyp category knowledge distillation (proposed) for which class aware weights are $\lambda_{ndbe} = 0.165$, $\lambda_{neoplasia} = 0.33$, $\lambda_{polyp} = 0.33$}
\label{table: KD results}
\begin{tabular}{lll|ccc}
\hline
\multicolumn{1}{l|}{\textbf{Method}}                                                             & \multicolumn{1}{l|}{\textbf{Augment}} & \textbf{Epochs} & \multicolumn{1}{l|}{\textbf{mAP$_{25}$}}       & \multicolumn{1}{l|}{\textbf{mAP$_{50}$}}            & \multicolumn{1}{l}{\textbf{mAP$_{25:75}$}}     \\ \hline
\multicolumn{1}{l|}{\textbf{\begin{tabular}[c]{@{}l@{}}Faster R-CNN\\ (ResNet-50)\end{tabular}}} & \multicolumn{1}{l|}{None}             & 60              & \multicolumn{1}{c|}{50.2}                      & \multicolumn{1}{c|}{40.3}                           & \textbf{37.9}                                  \\ \hline
\multicolumn{1}{l|}{\textbf{Proposed}}                                                           & \multicolumn{1}{l|}{None}             & 60              & \multicolumn{1}{c|}{\textbf{53.9}}             & \multicolumn{1}{c|}{\textbf{42.1}}                  & 37.8                                           \\ \hline
                                                                                                 &                                       &                 & \multicolumn{3}{c}{\textbf{Class wise average precision}}                                                                                             \\ \cline{4-6} 
\textbf{}                                                                                        & \textbf{}                             & \textbf{}       & \multicolumn{1}{l|}{\textbf{AP$_{50_{ndbe}}$}} & \multicolumn{1}{l|}{\textbf{AP$_{50_{neoplasia}}$}} & \multicolumn{1}{l}{\textbf{AP$_{50_{polyp}}$}} \\ \hline
\multicolumn{1}{l|}{\textbf{\begin{tabular}[c]{@{}l@{}}Faster R-CNN\\ (ResNet-50)\end{tabular}}} & \multicolumn{1}{l|}{None}             & 60              & \multicolumn{1}{c|}{65.6}                      & \multicolumn{1}{c|}{\textbf{42.2}}                  & 12.9                                           \\ \hline
\multicolumn{1}{l|}{\textbf{Proposed}}                                                           & \multicolumn{1}{l|}{None}             & 60              & \multicolumn{1}{c|}{\textbf{71.9}}             & \multicolumn{1}{c|}{37.3}                           & \textbf{16.9}                                  \\ \hline
\end{tabular}
\end{table}

\begin{table}[t!]
\centering
\caption{\textbf{Evaluation on unseen test data.} Faster R-CNN (ResNet50) without (SOTA) and with informed polyp class knowledge distillation (proposed, with class aware loss weights $\lambda_{ndbe} = 0.165$, $\lambda_{neoplasia} = 0.33$, $\lambda_{polyp} = 0.33$).}
\label{table: Results unseen}
\begin{tabular}{lll|ccc}
\hline
\multicolumn{1}{l|}{\textbf{Method}}                                                             & \multicolumn{1}{l|}{\textbf{Augmentation}} & \textbf{Epochs} & \multicolumn{1}{l|}{\textbf{mAP$_{25}$}}       & \multicolumn{1}{l|}{\textbf{mAP$_{50}$}}            & \multicolumn{1}{l}{\textbf{mAP$_{25:75}$}}     \\ \hline
\multicolumn{1}{l|}{\textbf{\begin{tabular}[c]{@{}l@{}}Faster R-CNN\\ (ResNet-50)\end{tabular}}} & \multicolumn{1}{l|}{None}                  & 60              & \multicolumn{1}{c|}{46.6}                      & \multicolumn{1}{c|}{38.9}                           & 34.6                                           \\ \hline
\multicolumn{1}{l|}{\textbf{Proposed}}                                                           & \multicolumn{1}{l|}{None}                  & 60              & \multicolumn{1}{c|}{\textbf{48.9}}             & \multicolumn{1}{c|}{\textbf{42.6}}                  & \textbf{36.9}                                  \\ \hline
                                                                                                 &                                            &                 & \multicolumn{3}{c}{\textbf{Class wise average precision}}                                                                                             \\ \cline{4-6} 
\textbf{}                                                                                        & \textbf{}                                  & \textbf{}       & \multicolumn{1}{l|}{\textbf{AP$_{50_{ndbe}}$}} & \multicolumn{1}{l|}{\textbf{AP$_{50_{neoplasia}}$}} & \multicolumn{1}{l}{\textbf{AP$_{50_{polyp}}$}} \\ \hline
\multicolumn{1}{l|}{\textbf{\begin{tabular}[c]{@{}l@{}}Faster R-CNN\\ (ResNet-50)\end{tabular}}} & \multicolumn{1}{l|}{None}                  & 60              & \multicolumn{1}{c|}{\textbf{47.9}}             & \multicolumn{1}{c|}{17.8}                           & 51.0                                           \\ \hline
\multicolumn{1}{l|}{\textbf{Proposed}}                                                           & \multicolumn{1}{l|}{None}                  & 60              & \multicolumn{1}{c|}{42.8}                      & \multicolumn{1}{c|}{\textbf{21.1}}                  & \textbf{63.7}                                  \\ \hline
\end{tabular}
\end{table}

\vspace{-2mm}
\subsubsection{Qualitative results}
\begin{figure}[t!]
    \centering
    \includegraphics[width=0.95\textwidth]{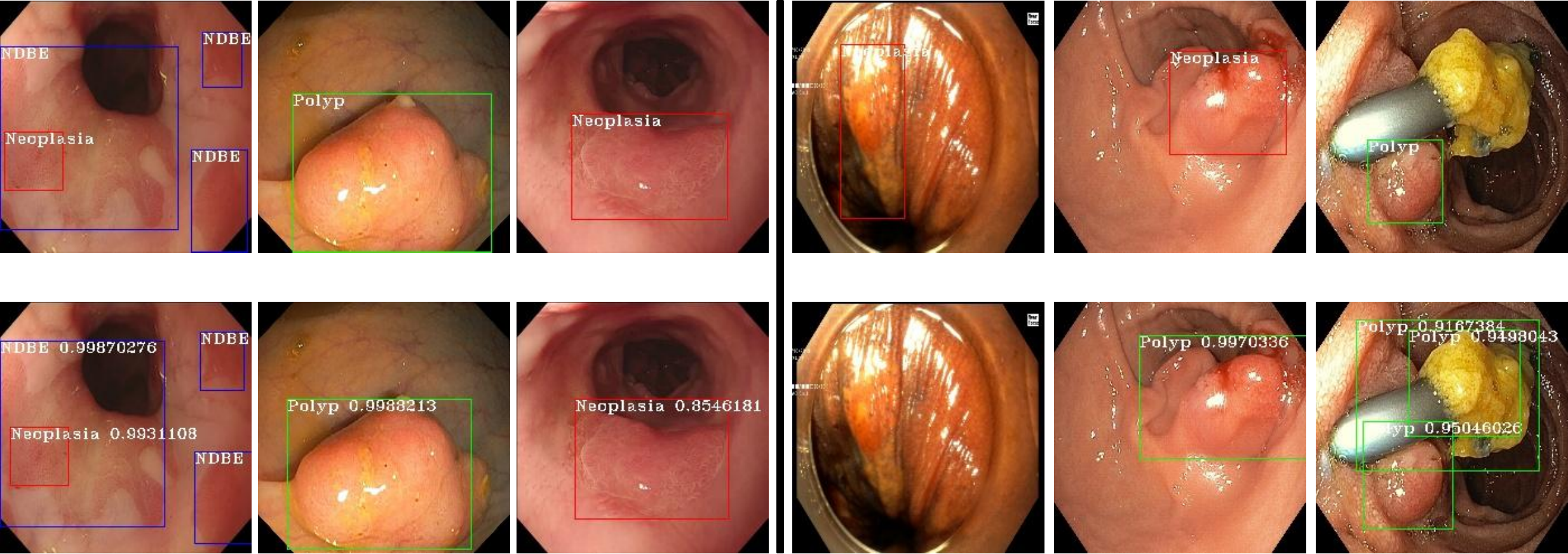}
    \caption{Qualitative results. First row shows images with their corresponding bounding boxes. Second row contains their corresponding predictions.}
    \label{fig: Results}
\end{figure}

Figure~\ref{fig: Results} presents a qualitative results on six samples of test set with their corresponding ground truth bounding boxes (on top) and detected boxes with classes from our proposed approach (on bottom). It can be observed that while protruded polyp class, semi-protruded neoplasia are detected precisely due to their appearance, however, there is a confusion with similar looking neoplastic changes (e.g., 5th column) or even other visual clutters (e.g., 6th column). Also, due to heterogeneous samples in neoplasia class (refer 4th column), the algorithm fails on under represented samples. It is also clear that in presence of multi-class (see 1st column), the algorithm is able to detect both non-dysplastic areas (in blue boxes) and early neoplasia (in red box).  

%
%
\vspace{-2mm}
\section{Conclusion}
\vspace{-2mm}
We have presented a class-aware knowledge distillation approach by leveraging class probability distances. The teacher branch is first trained using a larger dataset that is used to penalize the performance of the student model. Here, we proposed to use Bhattacharyya distance between class probability distributions. We demonstrate that our approach improves overall detection accuracies and can generalize well with the unseen test set, in particular, neoplasia and polyp that are clinically more important than non-dysplastic areas. In future work, we will use a third teacher model trained on another unique class.



%
\bibliographystyle{splncs04}
\bibliography{references}

\end{document}